\documentclass[letterpaper, 10 pt, conference]{ieeeconf}  

\IEEEoverridecommandlockouts                              

\usepackage{graphics} 
\usepackage{algorithm}
\usepackage[noend]{algpseudocode}

\usepackage{amsmath} 
\usepackage{amssymb}  
\usepackage{gensymb}

\usepackage{graphicx}
\usepackage{booktabs}
\usepackage{duckuments}
\usepackage{subfiles} 
\usepackage{acro}

\usepackage{caption}
\usepackage{subcaption}

\usepackage{cuted}
\usepackage{capt-of}

\usepackage{hyperref}
\hypersetup{colorlinks=true, citecolor=black, linkcolor=black, urlcolor=blue}

\DeclareAcronym{ugv}{
    short = UGV,
    long  = Unmanned Ground Vehicle,
    tag   = nomencl
}
\DeclareAcronym{uav}{
    short = UAV,
    long  = Unmanned Aerial Vehicle,
    tag   = nomencl
}
\DeclareAcronym{vio}{
    short = VIO,
    long  = Visual-Inertial Odometry,
    tag   = nomencl
}
\DeclareAcronym{ukf}{
    short = UKF,
    long  = Unscented Kalman Filter,
    tag   = nomencl
}
\DeclareAcronym{cvgl}{
    short = CVGL,
    long  = Cross-View Geo-localisation,
    tag   = nomencl
}
\DeclareAcronym{paper_name}{
    short = TACO,
    long  = Trajectory Aligning Cross-view Optimisation,
    tag   = nomencl
}
\DeclareAcronym{fov}{
    short = FOV,
    long  = Field-of-View,
    tag   = nomencl
}
\DeclareAcronym{fovs}{
    short = FOVs,
    long  = Fields-of-View,
    tag   = nomencl
}
\DeclareAcronym{bev}{
    short = BEV,
    long  = Birds-Eye-View,
    tag   = nomencl
}
\DeclareAcronym{gnss}{
    short = GNSS,
    long  = Global Navigation Satellite Systems,
    tag   = nomencl
}
\DeclareAcronym{sota}{
    short = SOTA,
    long  = state of the art,
    tag   = nomencl
}
\DeclareAcronym{ate}{
    short = ATE,
    long  = Absolute Trajectory Error,
    tag   = nomencl
}
\DeclareAcronym{ape}{
    short = APE,
    long  = Absolute Pose Estimation,
    tag   = nomencl
}
\DeclareAcronym{rpe}{
    short = RPE,
    long  = Relative Pose Estimation,
    tag   = nomencl
}
\DeclareAcronym{dof}{
    short = DoF,
    long  = Degrees of Freedom,
    tag   = nomencl
}
\DeclareAcronym{cdf}{
    short = CDF,
    long  = Cumulative Distribution Function,
    tag   = nomencl
}

\title{\LARGE \bf
TACO: Trajectory Aligning Cross-view Optimisation
}

\author{
Tavis Shore$^{1}$ and Oscar Mendez$^{1,2}$ and Simon Hadfield$^{1}$
\thanks{$^{1}$ Centre for Vision Speech and Signal Processing, University of Surrey,
Guildford, United Kingdom, \{\tt\small t.shore, s.hadfield\}@surrey.ac.uk}
\thanks{$^{2}$ Locus Robotics, Kent, United Kingdom, \tt\small omendez@locusrobotics.com}
}

\begin{document}
\maketitle

\setlength{\stripsep}{0pt plus 0pt minus 0pt}
\begin{strip}
\vspace*{-4\baselineskip}
\centering
\includegraphics[width=\textwidth]{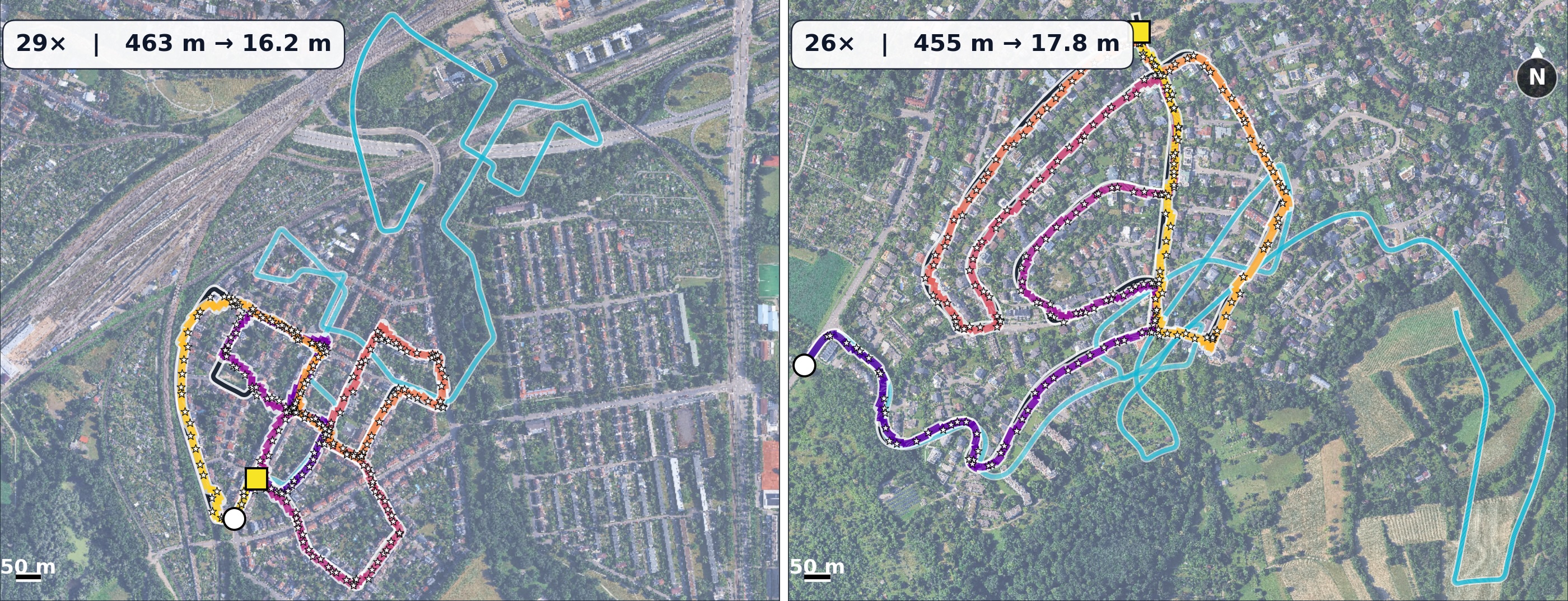}
\captionof{figure}{TACO trajectories closely track ground truth, whilst IMU-only (blue) drifts unboundedly.}
\label{fig:teaser}
\vspace*{0.5em}
\end{strip}

\thispagestyle{empty}
\pagestyle{empty}

\begin{abstract}
\ac{cvgl} matches ground imagery against satellite tiles to give absolute position fixes, an alternative to GNSS where signals are occluded, jammed, or spoofed. 
Recent fine-grained CVGL methods regress sub-tile metric pose, but have only been evaluated as one-shot localisers, never as the primary fix in a live pipeline. 
Inertial sensing provides high-rate relative motion, but accumulates unbounded drift without an absolute anchor.
We propose TACO, a tightly-coupled IMU + fine-grained \ac{cvgl} pipeline that consumes a single GNSS reading at start-up and thereafter operates on onboard sensing alone.
A closed-form cross-track error model triggers \ac{cvgl} before IMU drift exceeds the matcher's capture radius, and a forward-biased five-point multi-crop search keeps inference cost fixed at five forward passes per fix.
A yaw-residual gate rejects fixes that disagree with the onboard compass, and an anisotropic body-frame noise model scales each Unscented Kalman Filter update by per-fix confidence.
A factor graph with vetted loop closures provides an offline smoothed trajectory. 
On the KITTI raw dataset, TACO reduces median \ac{ate} from $\mathbf{97.0}$\,m (IMU-only) to $\mathbf{16.3}$\,m, a $\mathbf{5.9\times}$ reduction, at $\mathbf{<0.1}$\,ms per-frame fusion cost and a $\mathbf{5}$-$\mathbf{10\%}$ camera duty cycle. 
Code is available: \href{https://github.com/tavisshore/TACO}{github.com/tavisshore/TACO}.

Keywords: Localisation, Vision-Based Navigation, Sensor Fusion
\vspace{-0.5em}
\end{abstract}

\section{Introduction}
Localisation underpins autonomy in mobile robotics. Most deployed systems still rely on \ac{gnss} for an absolute reference, but satellite positioning fails exactly where autonomy is most demanded: urban canyons where signals are occluded and reflected, and contested airspace where adversaries jam or spoof the constellation. Robust autonomous platforms cannot depend on services they do not control. The open problem is absolute, drift-bounded positioning over indefinite runtimes using only the sensors carried on the vehicle, with no live external communication.

The dominant alternatives are \ac{vio} and visual SLAM. Both produce only relative pose: \ac{vio} achieves sub-metre short-term accuracy but accumulates unbounded drift, while SLAM bounds error only when the platform revisits previously mapped regions or is supplied with a pre-built map. Neither suits a first-traversal run in unmapped terrain.

\ac{cvgl} offers an absolute anchor by matching a ground-level image to publicly available satellite imagery, a reference that already covers the planet at sub-metre resolution. Early retrieval-based \ac{cvgl} produced tile-level estimates of tens to hundreds of metres, too coarse to fuse against an inertial estimate. Recent fine-grained methods~\cite{xia2023ccvpe,lentsch2023slicematch,xia2025fg2} regress sub-tile metric pose, bringing \ac{cvgl} accuracy into the regime of a few seconds of low-cost IMU dead-reckoning. This makes \ac{cvgl} viable as the absolute correction signal in a sensor-fusion loop. However, to date these methods have been evaluated as one-shot localisers, never as the primary fix in a live, continuously running pipeline on a moving vehicle.

We close this gap with \ac{paper_name}. An IMU dead-reckons body-frame motion at sensor rate; a closed-form cross-track error model tracks accumulated uncertainty and triggers a \ac{cvgl} update before drift exceeds the capture radius of the fine-grained matcher. A forward-biased five-point multi-crop search keeps the matcher inside that radius at a fixed cost of five forward passes per trigger. A yaw-residual gate rejects fixes that disagree with the onboard compass, and an anisotropic body-frame noise model scales each \ac{ukf} update by per-fix confidence. A factor graph with geometrically vetted loop closures provides an offline smoothed trajectory. The result is a drift-bounded, map-free, GNSS-free localiser that runs in real time on a single consumer GPU with the sensor suite already present on most outdoor autonomous platforms.

\noindent Our contributions are:
\begin{itemize}
    \item The first live localiser to use fine-grained \ac{cvgl} as the sole absolute anchor, replacing live GNSS with a single start-up fix and onboard sensing.
    \item An IMU-error-triggered, forward-biased five-point search that stays inside the \ac{cvgl} capture radius at fixed inference cost per trigger.
    \item A yaw-residual gate and anisotropic per-fix noise model that admit only geometrically consistent \ac{cvgl} fixes into the \ac{ukf}.
\end{itemize}

\section{Related Works}
\subsection{GNSS-denied Inertial-Visual Localisation}
Visual-inertial odometry systems such as VINS-Mono~\cite{qin2018vinsmono}, ORB-SLAM3~\cite{campos2021orbslam3}, and OKVIS~\cite{leutenegger2015okvis} achieve sub-metre short-term accuracy by tightly fusing IMU pre-integration with visual feature tracks, but produce only relative pose and accumulate unbounded drift on trajectories without revisits.
LiDAR-inertial systems such as LIO-SAM~\cite{shan2020liosam} and FAST-LIO2~\cite{xu2022fastlio2}, and HD-map matching approaches that localise a camera within a prior LiDAR map~\cite{wolcott2014visual}, bound error against a prior, but require either an expensive sensor or a pre-built map of the deployment area, neither available for first-traversal missions in unprepared terrain. We instead pair an IMU and monocular camera with freely available satellite imagery, the lightest sensor footprint that still provides an absolute reference without prior survey or infrastructure.

\subsection{Cross-View Geo-Localisation}
Retrieval-based \ac{cvgl} formulates ground-to-aerial matching as nearest-neighbour search in a learned embedding space. CVM-Net~\cite{hu2018cvmnet} pairs a Siamese backbone with NetVLAD pooling, SAFA~\cite{shi2019safa} introduces a polar transform with spatial-aware attention, and TransGeo~\cite{zhu2022transgeo} replaces the CNN with a transformer. BEV-CV~\cite{shore2024bevcv} lifts ground images into a semantic bird's-eye-view before matching, narrowing the cross-view domain gap by aligning representations rather than warping the aerial image. All report strong top-1 recall on CVUSA~\cite{workman2015localize} and VIGOR~\cite{zhu2021vigor}, but their output is a discrete tile selection: localisation error is dominated by tile size and typically falls in the tens of metres, an order of magnitude too coarse to anchor an IMU whose cross-track drift over a few seconds is already sub-metre.

Fine-grained \ac{cvgl} regresses metric pose within a tile, a direction catalysed by the VIGOR benchmark~\cite{zhu2021vigor} which exposed the inadequacy of tile-level metrics. CCVPE~\cite{xia2023ccvpe} predicts a 2D location heat-map with a conditional orientation field, SliceMatch~\cite{lentsch2023slicematch} aggregates aerial features along ground-frustum slices, Shi~\emph{et~al.}~\cite{shi2023boosting} iteratively refine 3-DoF pose with a geometry-guided cross-view transformer, and FG\textsuperscript{2}~\cite{xia2025fg2} matches ground-derived BEV points against aerial-sampled points and solves for pose by Procrustes alignment, the design we adopt as our matcher. These methods reach metre-level accuracy under good conditions, but all assume the query lies near the centre of a known aerial crop (equivalently, a coarse location prior of the order of the matcher's capture radius), and every published evaluation is one-shot: a single ground image against a single aerial crop, with no temporal context and no fusion. PEnG~\cite{peng} relaxes the known-crop assumption by cascading graph-based coarse retrieval with edge-localised \ac{rpe} over a city-scale road graph, but still operates one-shot per query and at a per-query cost incompatible with real-time deployment. To our knowledge, no fine-grained \ac{cvgl} method has yet been demonstrated as the absolute fix in a live, continuously running localiser.

\subsection{Fusing CVGL with Inertial Sensing}
A small body of work fuses \ac{cvgl} with onboard motion estimates. Shetty and Gao~\cite{shetty2019uav} fuse a Siamese cross-view network with monocular VO through a Kalman filter on a UAV. Dixit~\emph{et~al.}~\cite{dixit2020evaluation} treat retrieval-based cross-view matches as sensor measurements inside a particle filter on a ground vehicle. Jin~\emph{et~al.}~\cite{jin2024bevrender} (BEVRender) cast GNSS-denied UGV localisation as template matching of an inferred local BEV against a georeferenced aerial map. OrienterNet~\cite{sarlin2023orienternet} localises a single ground image against OpenStreetMap given a coarse GPS prior. None of these use fine-grained pose-regressing \ac{cvgl} as the primary absolute fix in a tightly-coupled IMU loop, and none demonstrates indefinite-runtime, drift-bounded operation on a moving ground vehicle without any live GNSS signal. \ac{paper_name} closes this gap by using the IMU's dead-reckoned trajectory both as the coarse prior the fine-grained matcher requires and as the high-rate state the matcher's fixes correct.

\section{Methodology}\label{sec:methodology}

\begin{figure*}[t!]
    \centering
    \includegraphics[width=\textwidth]{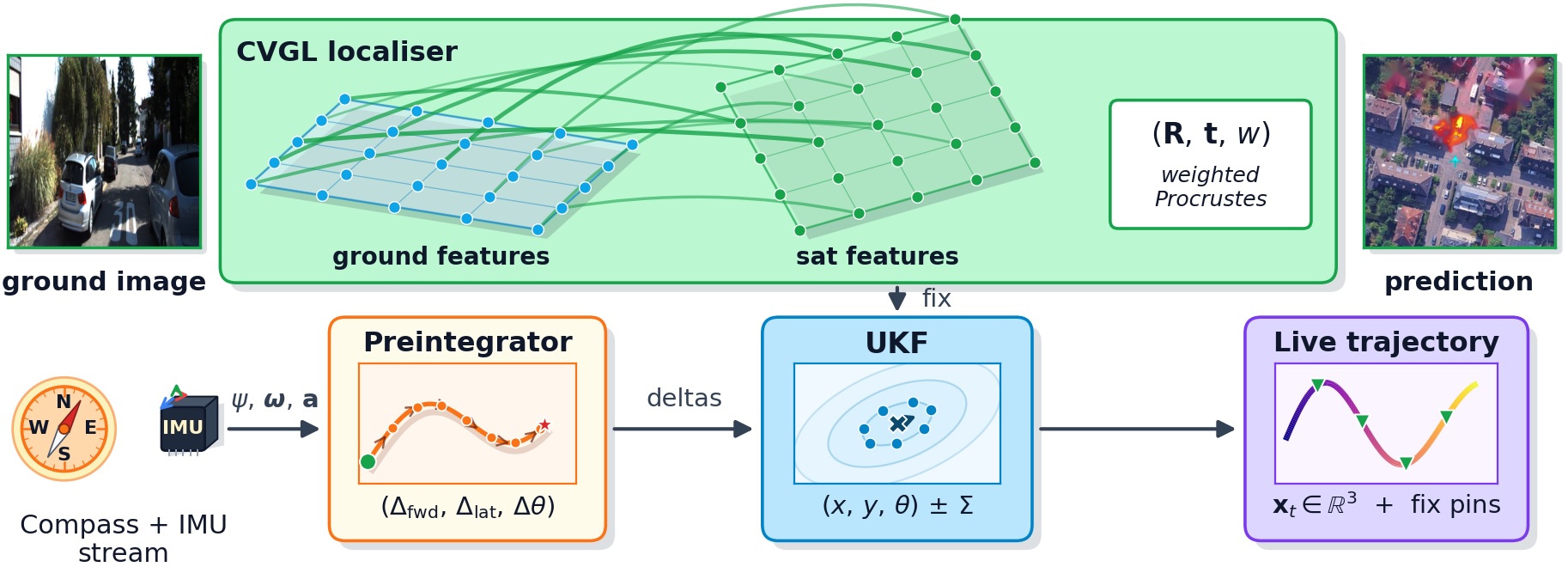}
    \caption{IMU stream feeds a preintegrator (reset per accepted fix); the IMU error trigger drives \ac{cvgl} inference on a forward-biased satellite crop, and the yaw-gated \ac{cvgl} fix updates a UKF whose accepted fixes anchor a Pose2 factor graph with vetted loop closures.}
    \label{fig:pipeline}
    \vspace{-1.75em}
\end{figure*}

\subsection{Problem Formulation and Notation}
We track a planar pose $x_t = [x_t, y_t, \theta_t]^\top$ in metres and radians relative to a UTM origin set from a single GNSS reading at start-up; no further GNSS is consumed.
An IMU sample $(\boldsymbol{\omega}_t, a_t, \Delta t_t)$ arrives at every step.
A ground-level RGB image $I_t$ with intrinsics $K_t$ and a corresponding crop of the globally referenced zoom-19 satellite mosaic $S$ are queried only when the trigger of Section~\ref{subsec:trigger} fires, on a small fraction of IMU steps.
Between triggers the camera can remain in a low-power standby state, so the visual sensor's average duty cycle scales with the trigger rate rather than with wall-clock time.
The objective is to produce $x_t$ online with \ac{ate} that remains bounded as the run length $T \to \infty$, without revisiting previously mapped regions, without a survey drive, and without live external communication.

\subsection{IMU Preintegration and Error Model}
We maintain a dead-reckoning position-error estimate since the last accepted fix from two complementary sources. The gyro-driven cross-track random walk and the accelerometer-driven along-track contribution are
\begin{equation}
\hat{\epsilon}_{\mathrm{cross}}(T) = 3\,d(T)\,\sigma_\omega\sqrt{T}, \qquad
\hat{\epsilon}_{\mathrm{along}}(T) = \tfrac{1}{2}\,\sigma_a(T)\,T^2,
\label{eq:imu_err}
\end{equation}
where $T$ is the elapsed time since the anchor, $d(T) = \sum_k v_k \Delta t_k$ is the distance travelled, $\sigma_\omega$ is the gyroscope angle random walk in $\mathrm{rad}/\sqrt{\mathrm{s}}$ calibrated to the platform's IMU, and $\sigma_a(T)$ is the empirical standard deviation of the high-pass-filtered forward accelerometer residual over the same window; the constant component is absorbed by the velocity estimate, leaving un-modelled bias drift as the relevant error source. Under the small-angle assumption, $d(T)\,\sigma_\omega\sqrt{T}$ has units of metres. The leading factor of $3$ on the cross-track term is a deliberate $3\sigma$ inflation: it fires the trigger before the true error reaches the matcher's capture radius with high probability, trading marginally higher fix density for a smaller probability of capture-radius overshoot.
The UKF heading state itself is supplied by an absolute heading measurement (the platform's compass-aided IMU), so~\eqref{eq:imu_err} drives only the trigger and query-bias predictor (Section~\ref{subsec:trigger}), never the realised heading evolution.

\paragraph{Preintegration.}
Body-frame motion between IMU samples is summarised by a preintegrator whose output drives both the UKF predict step and the factor-graph odometry chain. The preintegrator is reset on every accepted CVGL fix (details in Section~\ref{subsec:ukf}), keeping the trigger and the filter mutually consistent.

\paragraph{Composite trigger value.}
The scalar quantity used to gate inference and scale process noise is the envelope
\begin{equation}
\epsilon_{\mathrm{IMU}} = \max\!\left(\hat{\epsilon}_{\mathrm{cross}}(T),\ \hat{\epsilon}_{\mathrm{along}}(T),\ 0.03\,d_{\mathrm{div}}\right),
\end{equation}
where $d_{\mathrm{div}}$ is the chord between the current UKF position and the corrected preintegrator. The third term is a small floor that prevents $\epsilon_{\mathrm{IMU}}$ from collapsing to zero on benign segments where the model alone would underestimate accumulated uncertainty.

\subsection{Fine-Grained CVGL Module}
We use FG\textsuperscript{2}~\cite{xia2025fg2} as a black-box metric localiser. A query forward pass takes a ground image $I_t$ and a satellite crop $S_c$ centred on a candidate UTM coordinate, encodes both with a shared DINOv2 backbone, lifts ground features to a \ac{bev} grid, matches \ac{bev} cells against satellite cells via a learnt cross-view matcher, and returns a sub-pixel offset and rotation through weighted Procrustes alignment. The returned tuple is $(R, t, w)$: a $2{\times}2$ rotation, a metric translation in the rotated tile frame, and a scalar match weight $w \in (0,1]$ summarising the Procrustes mass.

The satellite source is a zoom-19 GeoTIFF mosaic at $\sim\!0.3\,\mathrm{m/px}$, cached locally and queried by lat/lon to UTM conversion. Each crop is rotated so that the vehicle heading aligns with image-up using $h_{\mathrm{img}} = \tfrac{\pi}{2} - \psi_{\mathrm{ENU}}$ to convert from the IMU's east-CCW yaw convention to the navigation convention expected by \ac{cvgl}. The network is used with its published weights, making our contribution backbone-agnostic: a gating, biasing, and noise-aware fusion architecture that turns sparse metric fixes into the absolute anchor of a real-time state estimator.

\subsection{Triggering and Multi-Crop Search}\label{subsec:trigger}

\paragraph{Dual trigger.}
A \ac{cvgl} inference is invoked when either the IMU error satisfies $\epsilon_{\mathrm{IMU}} \geq 1.0\,\mathrm{m}$, \emph{or} the wall-clock time since the last accepted fix exceeds $2.0\,\mathrm{s}$. The error trigger covers the moving regime, where drift is bounded by speed and gyro noise; the time trigger covers the stationary regime, where $\epsilon_{\mathrm{IMU}}\to 0$ so the model alone would never call for a fix and the filter would lose its anchor on long stops. Together they keep the fix rate roughly proportional to information demand: dense fixes during cornering and traffic, sparse fixes on smooth highway runs.

\paragraph{Forward bias.}
The UKF position at trigger time represents the vehicle at the \emph{last} anchor, not the vehicle now. Between anchor and trigger the vehicle has moved forward by approximately $\epsilon_{\mathrm{IMU}}$ along its current heading. We therefore shift the search centre forward by
\begin{equation}
\delta_{\mathrm{fwd}} = \min\!\left(0.4\,\epsilon_{\mathrm{IMU}},\ 15\,\mathrm{m}\right)
\end{equation}
in the heading direction before sampling crops. The conservative coefficient $0.4$ avoids overshoot when the vehicle is decelerating; the cap at $15\,\mathrm{m}$ avoids extrapolating off the satellite tile during long blackouts.

\begin{figure}[t!]
    \centering
    \includegraphics[width=\columnwidth]{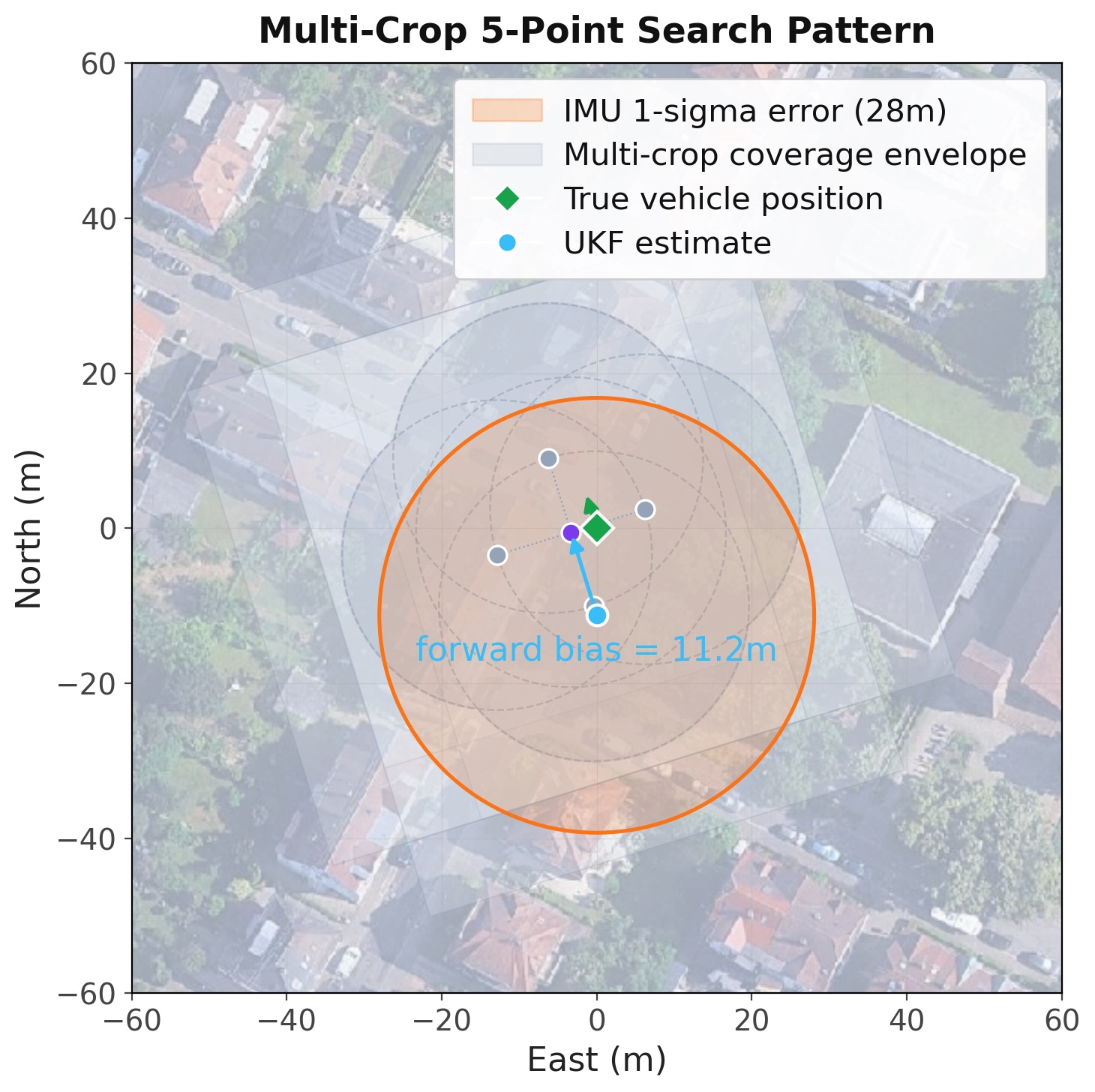}
    \caption{Multi-crop sampling at IMU error envelope $\epsilon_{\mathrm{IMU}}\!\approx\!28\,\mathrm{m}$: 5 body-frame offsets around the forward-biased UKF estimate; the candidate with the highest \ac{cvgl} weight wins.}
    \label{fig:multicrop}
    \vspace{-1.5em}
\end{figure}

\paragraph{Five-point cross.}
Around the bias-corrected centre we evaluate \ac{cvgl} at five body-frame offsets
\begin{equation}
\begin{gathered}
\{(\pm d, 0),\ (0, 0),\ (0, \pm d)\}, \\
d = \min\!\left(0.5\,\epsilon_{\mathrm{IMU}},\ 10\,\mathrm{m}\right),
\end{gathered}
\end{equation}
covering back/centre/forward/left/right. The candidate with the largest match weight $w$ is taken as the fix, and the inference cost per trigger is fixed at five forward passes. With FG\textsuperscript{2}'s empirical capture radius $r_{\mathrm{cap}}\approx 20\,\mathrm{m}$ at zoom-19, the union of the five capture discs covers a region of radius $\sim\!d + r_{\mathrm{cap}}$ around the search centre. At the trigger boundary ($\epsilon_{\mathrm{IMU}}=1\,\mathrm{m}$, $d=0.5\,\mathrm{m}$) the coverage envelope is $\sim\!20.5\,\mathrm{m}$, comfortably exceeding the residual offset between bias-corrected centre and true position. At the parameter caps ($d=10\,\mathrm{m}$, reached only after rejected-fix runs that drive $\epsilon_{\mathrm{IMU}}$ above $20\,\mathrm{m}$), the envelope grows to $\sim\!30\,\mathrm{m}$, providing headroom against the long-blackout regime.

\begin{figure}[t!]
    \centering
    \includegraphics[width=\columnwidth]{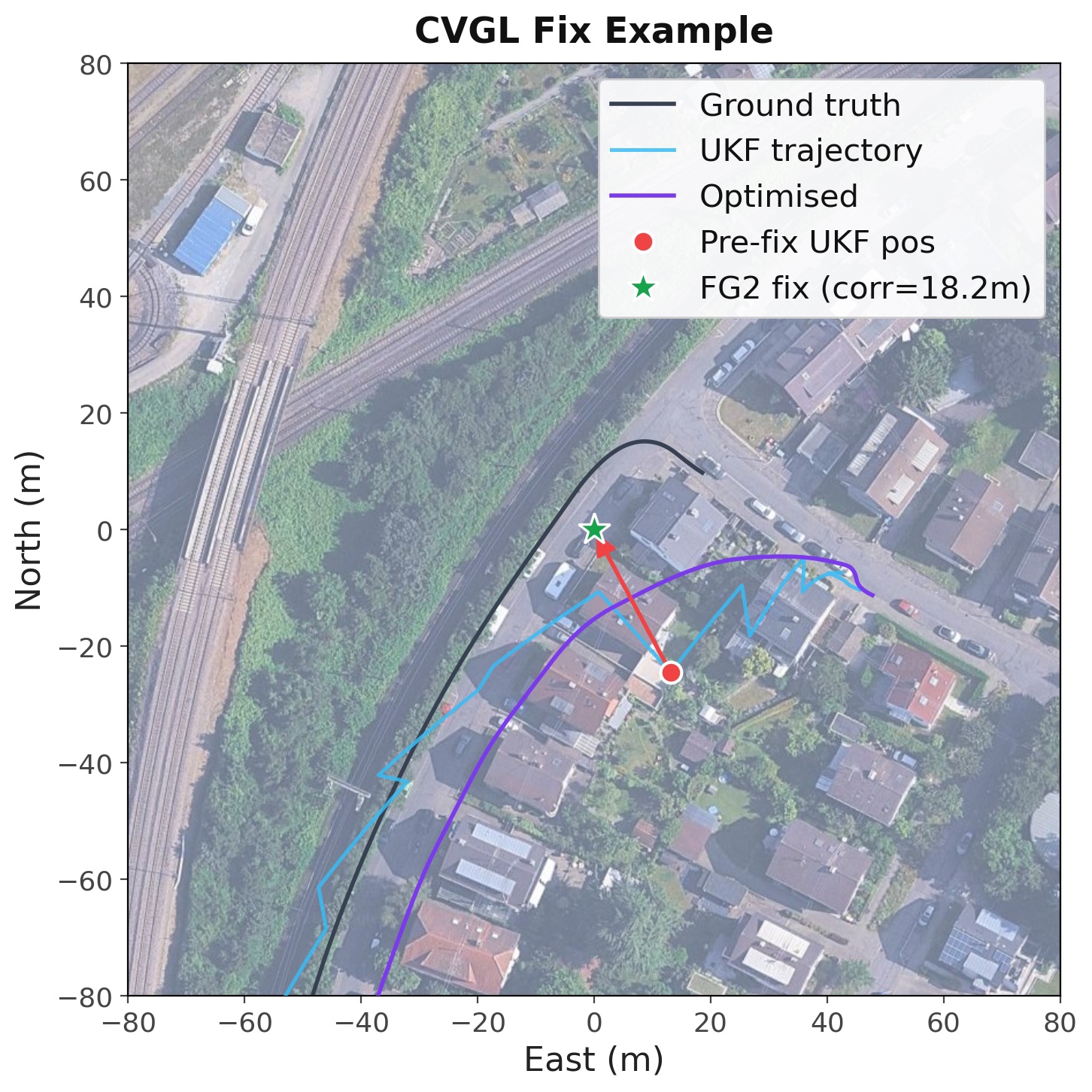}
    \caption{Single FG\textsuperscript{2} fix: the red arrow shows the UKF position correction at the green star, anchoring the drifting UKF (cyan) trajectory to the optimised path (purple).}
    \label{fig:fg2}
    \vspace{-1.5em}
\end{figure}

\subsection{Yaw-Residual Gate}\label{subsec:yawgate}
A surviving fix must pass a heading consistency check against the onboard compass before admission to the filter. Because the satellite tile is rotated to align with the compass-derived heading $\psi_{\mathrm{c}}$ before \ac{cvgl} is queried, the rotation matrix $R$ returned by the network encodes the residual between \ac{cvgl}'s predicted heading and $\psi_{\mathrm{c}}$:
\begin{equation}
    \psi_{\mathrm{res}} = \mathrm{atan2}\!\left(R_{1,0},\ R_{0,0}\right).
\end{equation}
We reject the fix if $|\psi_{\mathrm{res}}| > 0.35\,\mathrm{rad}\ (\approx 20^\circ)$. A large $\psi_{\mathrm{res}}$ indicates that the BEV matcher has locked onto a $90^\circ$- or $180^\circ$-symmetric structure (chequerboard intersections, parallel parking bays, motorway lane markings) and produced a high-$w$ but geometrically inconsistent correspondence; fusing the associated position fix would drag the filter away from truth. On rejection we update the time-since-fix counter as if a fix had been processed, which prevents the time trigger from re-firing immediately on the same bad scene.

\subsection{UKF Fusion with Anisotropic CVGL Noise}\label{subsec:ukf}

\paragraph{State and predict.}
The filter is a 3-state unscented Kalman filter on $(x, y, \theta)$ with standard van der Merwe weights ($\alpha = 10^{-3}$, $\beta = 2$, $\kappa = 0$). Sigma points are propagated through the body-frame rigid-motion model
\begin{equation}
\begin{aligned}
x' &= x + \cos\theta\,\Delta_{\mathrm{fwd}} - \sin\theta\,\Delta_{\mathrm{lat}}, \\
y' &= y + \sin\theta\,\Delta_{\mathrm{fwd}} + \cos\theta\,\Delta_{\mathrm{lat}}, \\
\theta' &= \theta + \Delta\theta,
\end{aligned}
\end{equation}
where $(\Delta_{\mathrm{fwd}}, \Delta_{\mathrm{lat}}, \Delta\theta)$ is the body-frame increment from the corrected preintegrator. The process noise is built directly from the IMU error model:
\begin{equation}
Q = \mathrm{diag}\!\left(\sigma_{\mathrm{fwd}}^2,\ \sigma_{\mathrm{lat}}^2,\ \sigma_{\theta}^2\right),
\end{equation}
with $\sigma_{\mathrm{fwd}} = \max(1.0,\ 2.0\,\epsilon_{\mathrm{IMU}})\,\mathrm{m}$, $\sigma_{\mathrm{lat}} = \max(1.5,\ 0.3\,\epsilon_{\mathrm{IMU}})\,\mathrm{m}$, and $\sigma_{\theta} = 0.05\,\mathrm{rad}$. The asymmetric scaling reflects that along-track odometry error grows linearly with speed while cross-track error is dominated by the gyro-noise floor.

\paragraph{Anisotropic position-only update.}
\ac{cvgl} returns a 2D position only, so the measurement model is $z = [x, y]^\top$. We construct the measurement covariance in the body frame and rotate it into the world frame using the heading at the localised frame:
\begin{equation}
R_{\mathrm{meas}} = \mathrm{Rot}(\theta)\,\mathrm{diag}\!\left(\sigma^2_{\mathrm{fwd},w},\ \sigma^2_{\mathrm{lat},w}\right)\mathrm{Rot}(\theta)^\top,
\end{equation}
where the body-frame sigmas are inverse-weight scaled,
\begin{equation}
\begin{aligned}
\sigma_{\mathrm{fwd},w} &= \mathrm{clip}\!\left(\tfrac{1.5}{w},\ 0.5,\ 8.0\right), \\
\sigma_{\mathrm{lat},w} &= \mathrm{clip}\!\left(\tfrac{3.0}{w},\ 1.0,\ 12.0\right).
\end{aligned}
\end{equation}
The factor of two between lateral and forward sigmas reflects FG\textsuperscript{2}'s BEV geometry: the matcher recovers along-heading position from rich foreshortened depth cues but resolves lateral position only through the much sparser cross-heading parallax. Inverse-$w$ scaling keeps weak fixes safe to fuse: a low-confidence return inflates $R_{\mathrm{meas}}$ to the cap, the Kalman gain shrinks, and the resulting state correction is bounded.

\paragraph{Position-only anchor reset.}
After a successful update we reset only the position channel of the corrected preintegrator to $(\hat x, \hat y)$ from the UKF; heading and velocity are preserved: heading is supplied by the compass at every step, and the IMU's short-horizon velocity estimate is unaffected by the position correction. The error model in~\eqref{eq:imu_err} resets at the same instant, so the trigger and the filter advance in lockstep.

\subsection{Factor Graph and Loop Closure}
The same data stream feeds a GTSAM Pose2 factor graph used for offline post-processing, enabling closer comparison to previous works that report smoothed trajectories. One $\mathrm{Pose2}$ node $x_i$ is instantiated per processed frame. Three factor families are added:
\begin{itemize}
    \item \emph{Origin prior} on $x_0$ at the start-up GNSS reading with isotropic noise $\sigma = 0.01$, treating it as effectively exact.
    \item \emph{Odometry} between consecutive nodes via $\mathrm{BetweenFactorPose2}$ with the body-frame increment from the corrected preintegrator and isotropic noise $\sigma_{\mathrm{odo},i} = \tfrac{1}{2}(\sigma_{\mathrm{fwd},i} + \sigma_{\mathrm{lat},i})$ taken from the same IMU error model that fed the UKF. A second, weaker $\mathrm{BetweenFactorPose2}$ with the lateral component zeroed and $\sigma = 1\,\mathrm{m}$ acts as a soft non-holonomic constraint that suppresses lateral jitter without preventing cornering.
    \item \emph{Fix priors} at every accepted CVGL frame, implemented as $\mathrm{PriorFactorPose2}$ with isotropic noise $\sigma_{\mathrm{fix},i} = \tfrac{1}{2}(\sigma_{\mathrm{fwd},w,i} + \sigma_{\mathrm{lat},w,i})$ taken directly from the per-fix sigmas used by the UKF measurement model, ensuring the graph and the filter agree on per-fix confidence.
\end{itemize}

A loop closure adds a relative-pose constraint between two fix nodes that observe the same physical location. To prevent spurious links between fixes that merely happen to be near one another (e.g.\ on parallel roads in a grid layout), we require three independent conditions to hold simultaneously for any candidate pair $(i, j)$: a UTM chord $d_{ij} \leq 50\,\mathrm{m}$ (a coarse spatial prune at the scale of the \ac{cvgl} coverage diameter at zoom-19); a frame gap $j - i \geq 30$ (preventing fixes within a single trajectory segment from being chained); and a path-to-chord ratio $\sum_{k=i}^{j-1}\lVert \Delta_k \rVert / d_{ij} \geq 5$. The path-to-chord test is the geometric discriminator: a genuine revisit forces the vehicle to detour substantially relative to the chord (around a block or after a U-turn), whereas fixes on a parallel road or in stop-and-go traffic yield ratios near $1$ and are rejected. A surviving pair contributes a $\mathrm{BetweenFactorPose2}$ with isotropic noise $\sigma_{\mathrm{lc}} = \max\!\bigl(\sqrt{\sigma_{\mathrm{fwd},w,i}^2 + \sigma_{\mathrm{fwd},w,j}^2},\ 0.5\,\mathrm{m}\bigr)$, so the loop edge is never more confident than the per-fix forward uncertainty at either endpoint, with a small absolute floor to prevent over-confident closures from dominating the optimisation. The full graph is solved by Levenberg-Marquardt for $100$ iterations at relative tolerance $10^{-5}$. The optimised easting and northing channels are then low-pass filtered with a Savitzky-Golay window of length $15$ and polynomial order $3$, removing residual node-level jitter while preserving curvature at junctions and turns.

\begin{figure*}[t!]
    \centering
    \includegraphics[width=\textwidth]{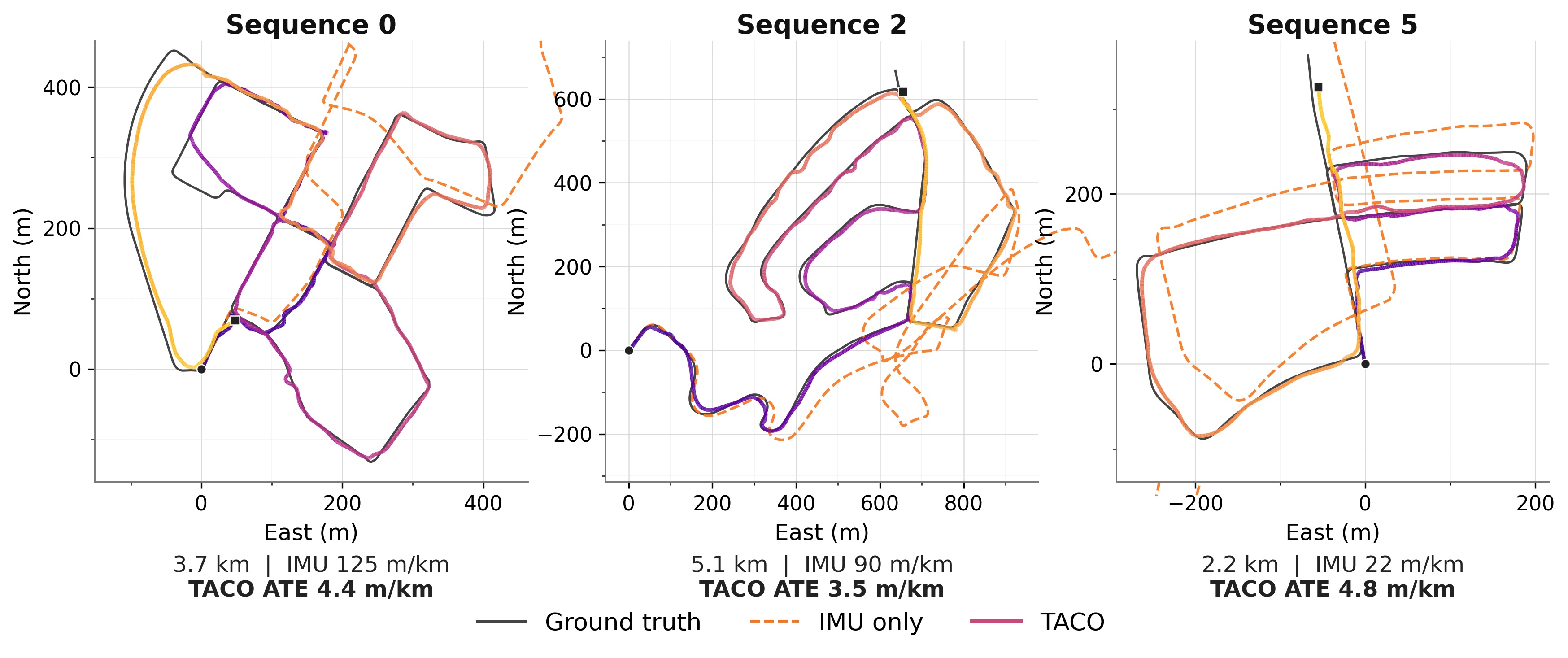}
    \caption{Trajectories on three KITTI sequences: IMU dead-reckoning drifts unboundedly while \ac{paper_name} tracks ground truth, demonstrating long-term bounded error from triggered \ac{cvgl} fixes.}
    \label{fig:traj}
\end{figure*}

\begin{table*}[h!]
\centering
\caption{Per-sequence ATE (m) on KITTI. Bold: best within minimal-sensor, real-time class (IMU, DROID-SLAM, DPVO, TACO). $^\star$ marks cross-area sequences disjoint from the FG\textsuperscript{2} training region. \ac{paper_name} entries are the UKF online trajectory.}
\label{tab:main}
\footnotesize
\setlength{\tabcolsep}{4.5pt}
\resizebox{0.9\textwidth}{!}{
\begin{tabular}{l rrrrrr c rr c}
\toprule
 & \multicolumn{6}{c}{\emph{Representative}} & \emph{Median} & \multicolumn{2}{c}{\emph{Non-rep.}} & \emph{Median} \\
\cmidrule(lr){2-7} \cmidrule(lr){9-10}
System & 00$^\star$ & 02 & 05 & 06$^\star$ & 08 & 10 & (rep.) & 01 & 04 & (non-rep.) \\
\midrule
IMU only & 463.4 & 455.4 & 47.7 & 66.6 & 127.4 & 22.5 & 97.0 & 41.6 & 1.4 & 21.5 \\
DROID-SLAM~\cite{teed2021droid} & 92.1 & \textit{fail} & 118.5 & 62.5 & 161.6 & 118.7 & 118.5 & 344.6 & 1.0 & 172.8 \\
DPVO~\cite{teed2023deep} & 113.2 & 123.4 & 59.0 & 54.8 & 115.9 & 13.6 & 86.1 & \textbf{12.7} & \textbf{0.7} & \textbf{6.7} \\
Fervers~\emph{et~al.}~\cite{fervers2022continuous} & - & 1.42 & 0.77 & 0.57 & 2.51 & 0.96 & 0.96 & 2.53 & 0.66 & 1.60 \\
Zhang~\emph{et~al.}~\cite{zhang2024increasing} & 0.95 & 0.82 & 0.54 & 1.31 & 1.17 & 0.65 & 0.89 & 1.52 & 0.36 & 0.94 \\
\textbf{TACO} & \textbf{16.2} & \textbf{16.4} & \textbf{10.5} & \textbf{17.0} & \textbf{83.1} & \textbf{12.9} & \textbf{16.3} & 93.4 & 5.2 & 49.3 \\
\midrule
IMU drift (m/km) & 124.8 & 90.1 & 21.7 & 54.3 & 39.7 & 24.5 & 47.0 & 17.0 & 3.5 & 10.2 \\
TACO ATE (m/km) & \textbf{4.4} & \textbf{3.5} & \textbf{4.8} & \textbf{13.8} & \textbf{25.9} & \textbf{14.8} & \textbf{9.3} & 38.3 & 13.2 & 25.8 \\
TACO steady-state RMSE (m) & 14.5 & 16.6 & 12.7 & 18.8 & 83.6 & 13.2 & 15.5 & 97.7 & 5.6 & 51.7 \\
TACO fixes/km & 62.7 & 45.9 & 61.8 & 43.2 & 65.4 & 62.1 & 62.0 & 23.7 & 28.1 & 25.9 \\
TACO smoothed (m) & 14.4 & 16.4 & 10.3 & 16.8 & 83.3 & 12.9 & 15.4 & 93.4 & 5.0 & 49.2 \\
\bottomrule
\end{tabular}
}
\vspace{-1.5em}
\end{table*}

\section{Results}
\subsection{Datasets and Protocol}
We evaluate on the KITTI odometry benchmark~\cite{kitti}, reporting all eight sequences with publicly available ground truth. Table~\ref{tab:main} splits them into a \emph{representative} group (multi-kilometre urban drives with turns, where gyro random-walk converts into measurable cross-track error) and a \emph{non-representative} group (near-straight highway, or too short for any inertial system to accumulate measurable drift). Group medians are reported separately so headline figures are not driven by sequences that do not test the bounded-drift claim. Sequences 00 and 06 are captured in regions disjoint from the FG\textsuperscript{2} training set, providing a cross-area generalisation test; the remaining sequences overlap the training region. Per-sequence cross-area status is marked in Table~\ref{tab:main}.

The IMU runs at $100\,\mathrm{Hz}$; camera frames are $10\,\mathrm{Hz}$ and consumed only at trigger times (Section~\ref{subsec:trigger}). At the dataset's mean velocity $8.2\,\mathrm{m/s}$ and the observed $\sim\!62$ accepted fixes/km on the representative group, this corresponds to a $\sim\!5\%$ camera duty cycle on the accepted-fix path; including yaw-rejected attempts pushes the upper end to $\sim\!10\%$. \ac{cvgl} inference is restricted to frames in the FG\textsuperscript{2} test splits~\cite{xia2025fg2} (with a short lookback to the most recent admissible split frame), eliminating training-set contamination. Satellite imagery is the zoom-19 Karlsruhe GeoTIFF cached locally. A single GPS reading at frame $0$ anchors the UTM origin; no GPS is consumed thereafter.

\subsection{Baselines and Metrics}
We compare against three classes of prior work, drawn from the published systems with both per-sequence KITTI ATE and compute information for a like-for-like comparison.

\paragraph{Monocular SLAM/VO.} DROID-SLAM~\cite{teed2021droid} and DPVO~\cite{teed2023deep} are learned monocular trajectory estimators; per-sequence values are reproduced from Lipson~\emph{et~al.}~\cite{lipson2024dpvslam} Table~2(b). Both apply Sim(3) alignment to ground truth, granting a scale freedom unavailable to \ac{paper_name}, which estimates absolute UTM coordinates. They share \ac{paper_name}'s monocular constraint but build their map online.

\paragraph{CVGL-anchored tracking.} Fervers~\emph{et~al.}~\cite{fervers2022continuous} and Zhang~\emph{et~al.}~\cite{zhang2024increasing} are the closest published architectures to \ac{paper_name}, each fusing \ac{cvgl} fixes with a continuous odometry source. Both invoke their \ac{cvgl} backbone on \emph{every} frame, and rely on richer sensor stacks: Fervers~\emph{et~al.} require LiDAR for ground-feature projection, Zhang~\emph{et~al.} use stereo cameras with a full ORB-SLAM3 front-end. The successor Fervers~\emph{et~al.}~\cite{fervers2023uncertainty} uses a four-camera surround rig and reports $2$--$3\,$Hz throughput on an RTX~6000, sub-real-time at KITTI's $10\,$Hz acquisition rate. 

\begin{table*}[tbp!]
\centering
\caption{Compute profile. Per-frame cost and GPU-s/$100\,$m at $10\,$Hz, $8.2\,$m/s.}
\label{tab:efficiency}
\footnotesize
\setlength{\tabcolsep}{5pt}
\resizebox{\textwidth}{!}{
\begin{tabular}{l l l l r r}
\toprule
\textbf{System} & \textbf{Sensors} & \textbf{Map} & \textbf{Hardware} & \textbf{Per-frame} & \textbf{GPU-s/100\,m} \\
\midrule
IMU only & IMU & --- & CPU & $\mathcal{O}(1)$ & $0$ \\
ORB-SLAM3~\cite{campos2021orbslam3} & mono cam & online & CPU & $50$~ms & --- \\
DROID-SLAM~\cite{teed2021droid} & mono cam & online & RTX 3090, $20\,$GB$^\dagger$ & $59$~ms & $7.2$ \\
DPVO~\cite{teed2023deep} & mono cam & online & RTX 3090, $4\,$GB$^\dagger$ & $21$~ms & $2.5$ \\
OrienterNet~\cite{sarlin2023orienternet} & mono cam & OSM & GPU & $88$~ms & $10.7$ \\
Fervers~\emph{et~al.}~\cite{fervers2022continuous} & mono cam + LiDAR + IMU & satellite & n.r. & n.r.$^\ddagger$ & n.r. \\
Fervers~\emph{et~al.}~\cite{fervers2023uncertainty} & 4$\times$surround cam + IMU & satellite & RTX 6000 & $\sim\!400$~ms$^\S$ & $\sim\!49$ \\
Zhang~\emph{et~al.}~\cite{zhang2024increasing} & stereo cam & satellite & RTX 3090 & $>$ ORB-SLAM3$^\P$ & $>10$\textsuperscript{est} \\
\textbf{TACO} & \textbf{mono cam + IMU} & \textbf{satellite} & \textbf{RTX 3090, 1.5GB} & $\mathbf{<\!1}$ / $\mathbf{336}$~\textbf{ms}$^\#$ & $\mathbf{2.0}$--$\mathbf{4.1}$ \\
\bottomrule
\end{tabular}
}
\vspace{2pt}
\begin{minipage}{\textwidth}
\footnotesize
$^\dagger$VRAM as reported on EuRoC~\cite{lipson2024dpvslam} Table~3; KITTI memory not separately reported. $^\ddagger$Per-frame ConvNeXt-T + UperNet with LiDAR projection; throughput not stated. $^\S$Authors report $2$-$3\,$Hz; per-frame cost is dataset-agnostic at $10\,$Hz. $^\P$Authors state pipeline ``requires more computational resources'' than ORB-SLAM3 alone. $^\#$IMU/UKF/error: $<\!1$~ms; CVGL: $\sim\!336$~ms per $5$-crop trigger ($\sim\!67$~ms/crop) on $5$-$10\%$ of frames. n.r.\ = not reported.
\end{minipage}
\end{table*}

\paragraph{OSM reference.} OrienterNet~\cite{sarlin2023orienternet} matches against rasterised OpenStreetMap tiles. We retain it as a per-frame compute reference but exclude it from the trajectory comparison: the prior modality differs, and no per-sequence KITTI ATE is published under a comparable protocol.

\paragraph{Metrics.} ATE is computed as RMSE in UTM metres after rigid first-pose alignment. Drift rate (m/km), steady-state RMSE (per-frame error after the third FG\textsuperscript{2} fix), and \ac{cvgl} fix density (per km) characterise \ac{paper_name}'s behaviour. Compute is reported as per-frame inference cost and GPU-seconds per $100\,$m at $8.2\,$m/s, the dataset's mean velocity.

\subsection{Quantitative Results}
\paragraph{Per-sequence ATE.}
\ac{paper_name} reduces median ATE from $97.0\,\mathrm{m}$ (IMU-only) to $16.3\,\mathrm{m}$ on the representative group (a $5.9\times$ reduction), and posts the lowest ATE in the minimal-sensor class on every representative sequence, recovering on seq~02 where DROID-SLAM~\cite{teed2021droid} fails.

\paragraph{CVGL-anchored systems.}
Fervers~\emph{et~al.}~\cite{fervers2022continuous} and Zhang~\emph{et~al.}~\cite{zhang2024increasing} achieve sub-metre to low-metre ATE; the gap is the price of their additional sensors and per-frame \ac{cvgl} invocation (Table~\ref{tab:efficiency}).

\paragraph{Non-representative group.}
On seq 01 and seq 04, IMU drift is already within \ac{cvgl}'s sub-fix uncertainty, so cross-view anchoring offers no benefit, consistent with framing \ac{paper_name} as a \emph{drift-bounded} replacement for GNSS rather than an accuracy-improving overlay on a low-drift baseline. DPVO posts the lowest minimal-sensor ATE on these sequences because Sim(3) alignment maps near-straight or short trajectories onto ground truth almost exactly, an advantage absolute-UTM \ac{paper_name} cannot exploit.

\paragraph{Compute footprint.}
Aggregating per-frame cost over the trigger schedule yields $2.0$-$4.1$ GPU-seconds per $100\,$m (Table~\ref{tab:efficiency}), an order of magnitude below every comparator. \ac{paper_name} uniquely combines a single-camera-and-IMU budget with bounded drift and real-time inference on a consumer GPU.

\subsection{Ablations}
Table~\ref{tab:ablation} isolates each design choice in Section~\ref{sec:methodology}. The yaw-residual gate, multi-crop search, and anisotropic measurement noise each contribute a comparable share of the headline reduction. Removing the forward bias hurts most on long sequences where the UKF anchor lags the live position by tens of metres at trigger time. The Pose2 smoother provides a further modest reduction on top of the UKF online trajectory, primarily by tightening curvature at loop closures.

\begin{table}[H]
\centering
\caption{Component ablation - representative group.}
\label{tab:ablation}
\setlength{\tabcolsep}{4pt}
\resizebox{0.85\columnwidth}{!}{
\begin{tabular}{l r r}
\toprule
    Configuration & ATE / km (m/km) \\
    \midrule
    \ac{paper_name} (full)              & \textbf{9.3} \\
    - no yaw-residual gate              & 13.0 \\
    - single crop                       & 16.6 \\
    - isotropic noise                   & 13.4 \\
    - no forward bias                   & 13.4 \\
    - no Pose2 smoother (UKF only)      & 10.4 \\
\bottomrule
\end{tabular}
}
\end{table}

\subsection{Drift Bounding over Time}
\begin{figure}[H]
    \centering
    \includegraphics[width=\columnwidth]{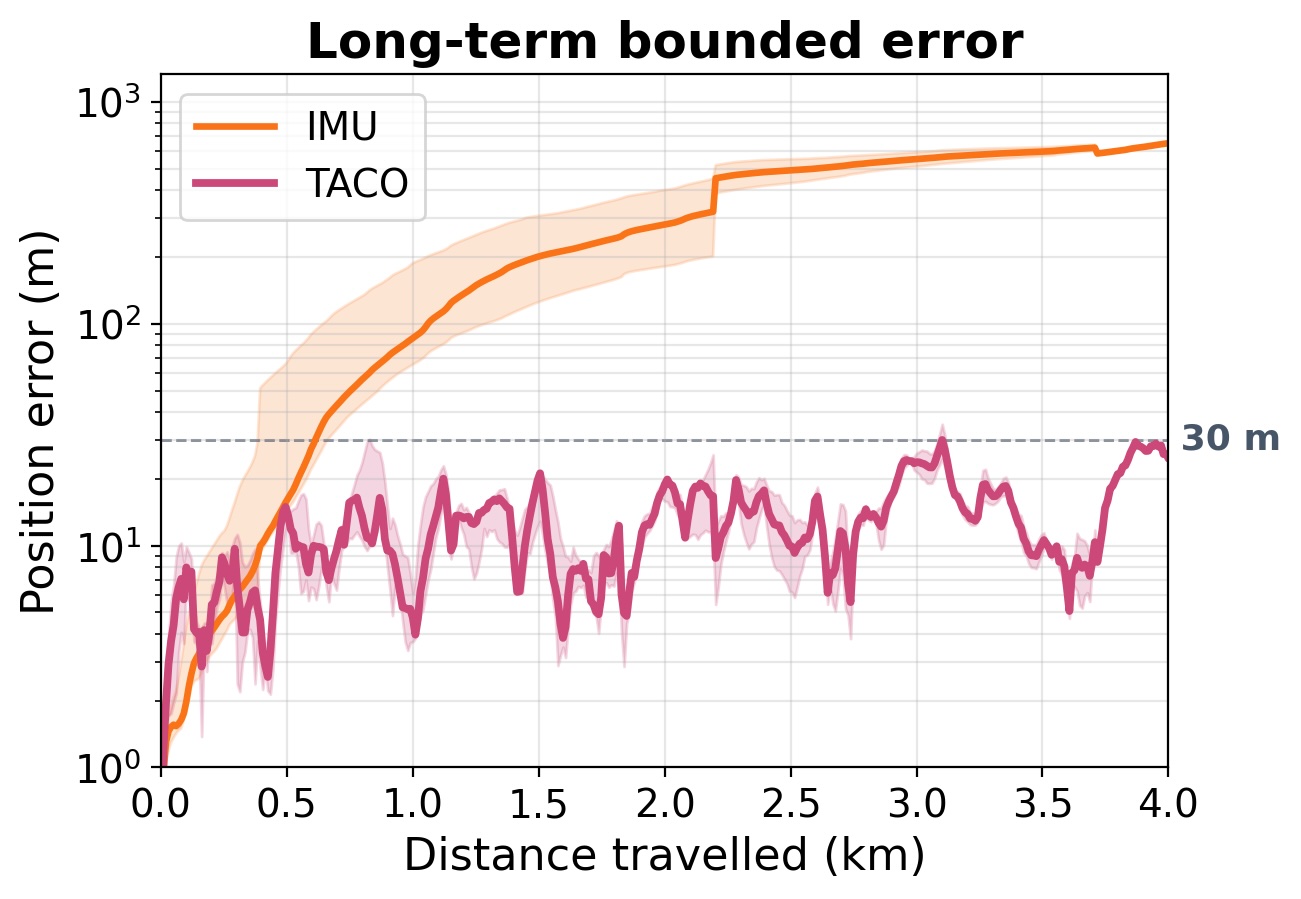}
    \caption{Median position error vs distance across sequences. IMU drifts unboundedly; \ac{paper_name} stays bounded near $30\,$m.}
    \label{fig:drift}
\end{figure}

Figure~\ref{fig:drift} aggregates per-frame position error against cumulative distance across representative sequences. IMU-only error grows super-linearly with distance, consistent with the $O(d\sqrt{t})$ random-walk model of Section~\ref{sec:methodology}. The fused \ac{paper_name} trajectory remains bounded around the $30\,\mathrm{m}$ reference envelope for the full multi-kilometre range, with momentary peaks at trigger boundaries and rapid relaxation after each accepted fix. Critically, the bound is set by the \ac{cvgl} capture radius and the trigger threshold, not by $T$: the empirical evidence behind the indefinite-runtime claim.

\subsection{Runtime}
On an NVIDIA RTX 3090 the IMU/UKF path runs at $<\!2\,$ms total per frame (preintegrator $<\!1\,$ms, UKF predict+update $\sim\!0.2\,$ms, error estimate $<\!1\,$ms). The \ac{cvgl} forward pass is $\sim\!67\,$ms per crop on our optimised inference path (DINO extractor in fp16, matcher under fp16 autocast, single ground feature reused across the candidate crops, satellite branch batched $B\!=\!5$); the multi-crop search therefore costs $\sim\!336\,$ms per trigger, amortising to $\sim\!17$-$34\,$ms per camera-rate frame at the $5$-$10\%$ trigger fraction. Peak VRAM is $\sim\!1.5\,$GB, verified to run under a hard $2\,$GB cap; this fits comfortably on Jetson Orin NX/Orin Nano-class modules, where TensorRT-FP16 deployment is expected to retain real-time operation. The low trigger fraction additionally enables camera standby between fixes, a power profile unavailable to the every-frame backbones of~\cite{fervers2022continuous,zhang2024increasing}.

\section{Conclusion \& Future Work}
\label{sec:future_work}
\ac{paper_name} demonstrates that fine-grained \ac{cvgl}, used previously only as a one-shot localiser, can serve as the sole absolute anchor in a live, indefinite-runtime localiser on a moving ground vehicle. On the KITTI representative group it reduces median ATE from $97.0\,\mathrm{m}$ (IMU-only) to $16.3\,\mathrm{m}$, a $5.9\times$ reduction, at a sustained per-frame fusion cost of $<0.1\,\mathrm{ms}$ and a \ac{cvgl} duty cycle of $5$-$10\%$, well over an order of magnitude below the GPU footprint of comparable map-aided localisers. 
Bounded drift, sub-millisecond fusion cost, and a backbone-agnostic design make \ac{paper_name} a viable option for autonomous platforms in GNSS-denied environments.

\subsection*{Future Work}
Further ATE reductions are gated by the \ac{cvgl} backbone rather than the fusion architecture: widening FG\textsuperscript{2}'s capture radius via offset-augmented retraining or a coarse-to-fine head would directly tighten the seq~08 ceiling and reduce the seq~01/04 regressions on sparse-fix segments. Porting \ac{paper_name} to aerial platforms introduces 6-DoF motion, faster yaw rates, and altitude-dependent scale variation, requiring retraining on overhead-oblique imagery and extending the UKF state to full SE(3). Finally, embedded deployment via backbone quantisation, distillation, or TensorRT compilation onto Jetson Orin Nano- or Raspberry Pi 5 NPU-class hardware would broaden applicability without architectural changes.



{\small
\bibliography{egbib.bib}
}

\end{document}